\documentclass[accepted]{melba}

\usepackage{mwe} 

\usepackage{amsmath,amsfonts}
\usepackage{enumitem}

\newcommand{\Diff}[0]{\mathrm{Diff}}    

\usepackage{xcolor}

\usepackage{amssymb}  
\usepackage{bm}       
\usepackage{float}
\usepackage{wrapfig}
\usepackage{makecell} 
\usepackage{multirow}
\usepackage{array}
\usepackage{stfloats}      

\newcolumntype{C}[1]{>{\centering\arraybackslash}m{#1}}

\newcommand{\tallerow}{\rule{0pt}{4.2ex}}

\melbaid{2025:019}  
\doi{10.59275/j.melba.2025-f9f4}  

\melbaauthors{Nian Wu and Miaomiao Zhang}  
\email{bsw3ac@virginia.edu}
\volume{3}
\firstpageno{425}  
\melbayear{2025}  
\datesubmitted{2024-10}  
\datepublished{2025-12}  

\ShortHeadings{Learning Geodesics of Geometric Shape Deformations}{Wu and Zhang}

\title{Learning Geodesics of Geometric Shape Deformations From Images}

\author{
	\firstname Nian \surname Wu\aff{1}\orcid{https://orcid.org/0000-0002-9168-3518},
	\name Miaomiao \surname Zhang\aff{1,2}\orcid{https://orcid.org/0000-0003-0457-3335}
}

\affiliations{
	\num 1 \addr University of Virginia, Department of Electrical \& Computer Engineering, Charlottesville, VA, USA \\
	\num 2 \addr University of Virginia, Department of Computer Science, Charlottesville, VA, USA \\
}

\abstract{
	This paper presents a novel method, named geodesic deformable networks (GDN), that for the first time enables the learning of geodesic flows of deformation fields derived from images. In particular, the capability of our proposed GDN being able to predict geodesics is important for quantifying and comparing deformable shape presented in images. The geodesic deformations, also known as optimal transformations that align pairwise images, are often parameterized by a time sequence of smooth vector fields governed by nonlinear differential equations. A bountiful literature has been focusing on learning the initial conditions (e.g., initial velocity fields) based on registration networks. However, the definition of geodesics central to deformation-based shape analysis is blind to the networks. To address this problem, we carefully develop an efficient neural operator to treat the geodesics as unknown mapping functions learned from the latent deformation spaces. A composition of integral operators and smooth activation functions is then formulated to effectively approximate such mappings. In contrast to previous works, our GDN jointly optimizes a newly defined geodesic loss, which adds additional benefits to promote the network regularizability and generalizability. We demonstrate the effectiveness of GDN on both 2D synthetic data and 3D real brain magnetic resonance imaging (MRI). 
    Our code is publicly available at ~\url{https://github.com/nellie689/GDN}.}


\begin{document}

\twocolumn[\maketitle]

\section{Introduction}
\label{sec:intro}

Deformable shape provides important features of objects presented in images, for example, abnormal shape changes of anatomical brain structures are shown to be predictors of neurodegenerative disorders~\citep{qiu2008parallel,lorenzi20104d,wang2007large,wang2022geo}. Existing methods have studied a variety of shape representations, including deformation-based descriptors that focus on highly detailed geometric information on dense images~\citep{christensen1993deformable,rueckert2003automatic}. With the underlying assumption that objects in many generic classes can be described as deformed versions of an ideal template, descriptors in this class arise naturally by aligning the template to an input image. The resulting transformation between images is then considered as a shape descriptor that reflects local geometric changes. A rich body of literature has been dedicated to harnessing such shape information for enhanced performance in image analysis, particularly in specialized domains of medical image analysis~\citep{bao20193d,beekman2022improving,sun2022topology} and computational anatomy~\citep{wang2007large,qiu2008parallel,grenander1998computational}. The realm of related research involves a suite of tools designed for comparing, matching, and modeling shapes based on distance metrics in deformation spaces~\citep{qiu2008parallel,hong2017fast,niethammer2019metric,qiu2012principal}. A {\em geodesic} (a.k.a. a curve whose tangent vector is parallel transported along itself) locally minimizes the Riemannian distance between points on a manifold. Such geodesics can quantify the geometric similarity between objects and serve as a solid mathematical foundation for deformable shape analysis, including regression~\citep{hong2017fast,niethammer2011geodesic}, longitudinal analysis~\citep{singh2013hierarchical,hong2019hierarchical}, and group comparisons~\citep{miller2004computational,qiu2008parallel}. Such a metric was first developed in the large deformation diffeomorphic metric mapping (LDDMM) algorithm~\citep{beg2005computing,vialard2012}, giving rise to a variational principle that expresses the optimal deformation as a geodesic flow of diffeomorphisms (i.e., bijective, smooth, and invertible smooth mappings). The geodesic path of deforming one image to another also serves to regularize the smoothness of the transformations grids with an intact topology of objects.

Recent advances in deep registration networks have achieved remarkable success in learning diffeomorphic transformations between images~\citep{yang2017quicksilver,balakrishnan2019voxelmorph,Wang_2020_CVPR,chen2022transmorph,kim2022diffusemorph}. These learning-based approaches characterize deformation fields using a time sequence of smooth vector fields governed by nonlinear differential equations. In contrast to traditional registration methods such as LDDMM that rely on iterative optimization schemes~\citep{beg2005computing,vialard2012}, learning-based methods offer rapid predictions without the need for adhoc manual parameter tuning during the testing phase. Despite the aforementioned advantages, current registration networks primarily focus on learning the initial conditions of deformation models, i.e., initial velocity fields~\citep{yang2017quicksilver,Wang_2020_CVPR,hinkle2018diffeomorphic}. The final deformation field is then obtained by solving a set of differential equations, which can be computationally expensive in high-dimensional image space during testing. More importantly, the definition of geodesic metrics in the deformation spaces, which is critical for quantifying and analyzing shape changes, is not explicitly considered or overlooked in the network learning process. This limits the interpretability and regularizability of the transformation process. Additionally, it may negatively impact the model generalizability by elevating the risk of overfitting the registration loss during training phase. 


Recent research on neural operators have been explored to learn the solution of ordinary /partial differential equations (O/PDEs)~\citep{lu2019deeponet,bhattacharya2021model,li2020neural,li2020fourier}. Among these, a notable work is the Fourier neural operator (FNO)~\citep{li2020fourier}, which achieves an inference time orders of magnitude faster than traditional numerical solutions to the fluid flow Navier-Stokes equations~\citep{constantin1988navier}, without sacrificing accuracy. Inspired by the work of~\citep{li2020fourier}, a research group had an initial attempt to utilize Fourier neural operators as a surrogate to learn the solution to geodesic flows governed by the well-known Euler-Poincaré differential equation (EPDiff)~\citep{arnold1966,miller2006} in the context of optimization-based image registration~\citep{wu2023neurepdiff}. Similarly, a group of research work formulate image registration as learning a transformation flow between images, where the deformation trajectories are obtained by integrating time-dependent velocity fields approximated by neural networks within a neural ODE framework~\citep{sun2024medical,wu2022nodeo}. However, they still need a conventional iterative optimization-based scheme for estimating the initial velocity fields during the registration inference phase.

In this paper, we introduce a novel geodesic deformable network (GDN) that {\em for the first time learns the geodesic metrics of latent deformation models directly from training images} in the context of registration networks. Our method builds upon FNO~\cite{li2023fourier} to model geodesic learning in a latent space, enabling efficient approximation of complex deformation dynamics. Unlike existing approaches~\citep{sun2024medical,wu2022nodeo}, our proposed method takes a fundamentally different approach by leveraging the EPDiff geodesic shooting equation to guide the learning of time-dependent velocity fields constrained along geodesic paths.
In comparison to the recent NeurEPDiff method~\citep{wu2023neurepdiff}, our model differs by treating the geodesics as unknown mapping functions in the latent deformation spaces, which are explicitly learned in the context of deep registration networks, without requiring optimization during the registration inference stage. A composition of integral operators and smooth activation functions is then formulated to effectively approximate such mappings in each hidden layer of the proposed GDNs. Our network will jointly optimize a newly defined geodesic loss, guided by the underlying transformation process solved through numerical integrators~\citep{vialard2012,beg2005computing}. A major benefit of the learned geodesic mapping function lies in its application as a computationally efficient surrogate model for the original numerical solutions of differential equations required to generate final deformations. In contrast to previous work~\citep{balakrishnan2019voxelmorph,Wang_2020_CVPR,chen2022transmorph,kim2022diffusemorph}, the advantages of our proposed method, GDN, are threefold:
\begin{itemize}
\item Predict a geodesic flow of deformation fields using a surrogate model, which is important for deep nets to quantify and analyze deformation-based shape objects.  
\item Promote the network regualrizability and generalizability via enforced geodesic constraints in deformation spaces.
\item Eliminate the need for numerical solutions of differential equations for final deformations in testing inference, a strategy that can be computationally advantageous in high-dimensional image spaces.
\end{itemize}


Experiments on 2D synthetic data and 3D real brain MRIs show that GDN is able to predict fairly close geodesics compared to conventional numerical solutions. We also demonstrate that GDN achieves superior performance in regularizability and generalizability when compared to the state-of-the-art deep learning-based registration networks in out-of-distribution (OOD) data~\citep{balakrishnan2019voxelmorph,hinkle2018diffeomorphic,chen2022transmorph,yang2017quicksilver}. 

\section{Background: Geodesics In Deformation Spaces}

\label{sec:backgroundlddmm}
In this section, we first briefly review the basic mathematical concepts of {\em geodesics} in the space of diffeomorphic transformations (a.k.a. diffeomorphisms). We then show how to derive the geodesic path of transformations from images using the LDDMM algorithm~\citep{beg2005computing} with the geodesic shooting equation~\citep{vialard2012,younes2009evolutions}.

\subsection{Geodesics of Diffeomorphisms}
Let $\Diff^\infty(\Omega)$ denote the space of smooth
diffeomorphisms on an image domain $\Omega$. The tangent space of diffeomorphisms is the
space $V = \mathfrak{X}^\infty(T\Omega)$ of smooth vector fields on $\Omega$. Consider a time-varying velocity field, $\{v_t\} : [0,\tau] \rightarrow V$, we can then generate diffeomorphisms $\{\phi_{t}\}$ as a solution to the equation
\begin{equation}
\label{eq:phi_v}
\frac{d \phi_t}{dt} = v_t(\phi_t), \, t \in [0, \tau].
\end{equation}

The LDDMM~\citep{beg2005computing} provides a \textit{distance metric} in the space of diffeomorphisms, which is also used as a regularization for image registration. Such a distance metric is formulated as an integral of the Sobolev norm of the time-dependent velocity field $v_t$, i.e., $\int_0^\tau (\mathcal{L} v_t, v_t) \, dt$, where $\mathcal{L}: V\rightarrow V^{*}$ is a symmetric, positive-definite differential operator that maps a tangent vector $ v(t)\in V$ into its dual space as a momentum vector $m(t) \in V^*$. We typically write $m(t) = \mathcal{L} v(t)$, or $v(t) = \mathcal{K} m(t)$, with $\mathcal{K}$ being an inverse operator of $\mathcal{L}$. In this paper, we adopt a commonly used Laplacian operator $\mathcal{L}=(- \alpha \Delta + \beta \, \text{Id})^2$, where $\alpha$ and $\beta$ are weighting parameters that control the smoothness of transformation fields and $\text{Id}$ is an identity matrix. The $(\cdot, \cdot)$ is a dual paring, which is similar to an inner product between vectors.

\begin{figure*}[!t]
\centering
\includegraphics[width=1.0\textwidth] {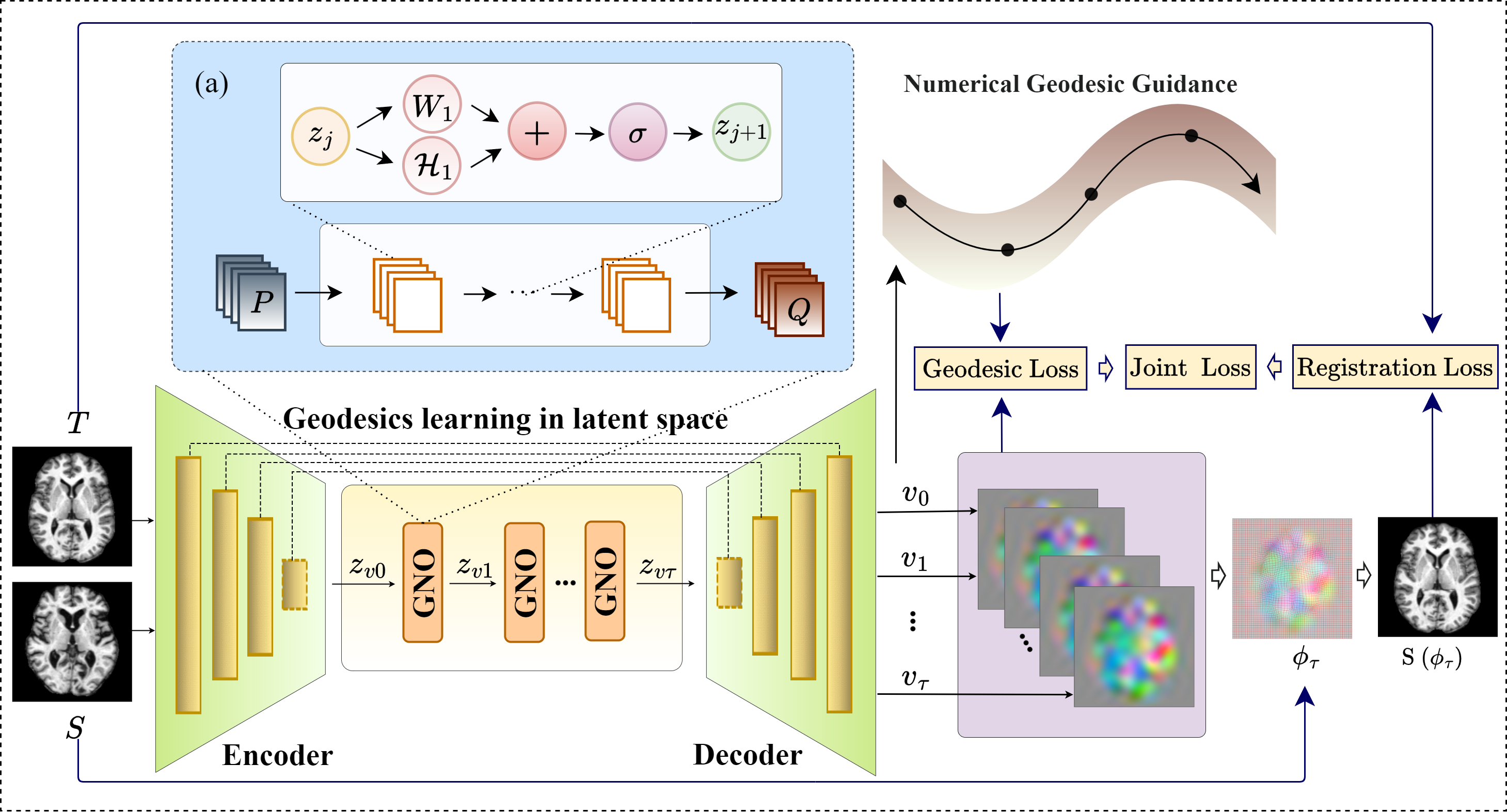}
     \caption{An overview of our proposed network GDN.}
\label{fig:NetArch}
\end{figure*}

According to a well-known geodesic shooting algorithm ~\citep{vialard2012}, the minimum of the previously defined distance metric above is uniquely determined by solving an Euler-Poincar\'{e} differential (EPDiff) equation~\citep{vialard2012,younes2009evolutions} with a given initial condition. That is, for $\forall v_0 \in V$, at $t
= 0$, a geodesic path $t \mapsto \phi_t \in \Diff^\infty(\Omega)$ in the space of diffeomorphisms can be computed by forward shooting the EPDiff equation formulated as
\begin{equation}
\label{eq:epdiff}
    \frac{\partial v_t}{\partial t} =-K\left[(Dv_t)^Tm_t + Dm_t\, v_t + m_t \operatorname{div} v_t\right],
\end{equation}
where $D$ denotes the Jacobian matrix and $\operatorname{div}$ is a divergence operator.

\subsection{LDDMM To Derive Geodesics From Images}
Assume a source image $S$ and a target image $T$ defined on a $d$-dimensional torus domain $\Omega = \mathbb{R}^d / \mathbb{Z}^d$ ($S(x), T(x):\Omega \rightarrow \mathbb{R}$). 
The space of diffeomorphisms is denoted by $\Diff(\Omega)$.
The problem of diffeomorphic image registration is to find the geodesic, to generate time-varying diffeomorphisms $\{\phi_t\}: t \in [0,\tau] $, such that a deformed source image by the smooth mapping $\phi_\tau$, noted as $S(\phi_\tau)$, is similar to $T$. By parameterizing the transformation $\phi_\tau$ with an initial velocity field $v_0$, we write the optimization problem of diffeomorphic registration in the setting of LDDMM with geodesic shooting as
\begin{align}
\label{eq:lddmm}
 E(v_0) = (\mathcal{L} v_0, v_0)  + \lambda  \text{Dist}(S(\phi_\tau), T )  \nonumber \\
 \text{s.t. Eq.}~\eqref{eq:phi_v}  \, \& \, ~\eqref{eq:epdiff}.
\end{align}
Here, Dist(·,·) is a distance function that measures the dissimilarity between images and $\lambda$ is a positive weighting parameter. The commonly used distance functions include the sum-of-squared intensity differences ($L_2$-norm)~\citep{beg2005computing}, normalized cross correlation (NCC)~\citep{avants2008symmetric}, and mutual information (MI)~\citep{wells1996multi}. In this paper, we use the sum-of-squared intensity differences. 

\section{Our Method}
\label{Sec:archNEPdiff}

In this section, we present a novel method, GDN, that learns geodesics of deformations in the context of deep registration networks. The key component of GDN is a newly designed geodesic neural operator (GNO) that encodes the underlying geodesic mapping functions in the latent feature space. A joint optimization of a typical unsupervised registration loss with a geodesic loss guided by numerical solutions of geodesic shooting equations will be introduced in GDN. An overview of our network GDN is shown in Fig.~\ref{fig:NetArch}.


\subsection{Geodesic Neural Operator}
Inspired by recent works on neural operators~\citep{kovachki2021neural,li2020fourier}, we develop GNO, noted by $\mathcal{G}_{\theta_r}$ with parameters $\theta_r$, as a surrogate model to approximate the solution to the geodesic shooting equations (Eq.~\eqref{eq:epdiff}) in the latent deformation spaces, parameterized by initial velocity fields.

We first employ a widely adopted unsupervised registration architecture featuring a UNet backbone~\citep{balakrishnan2019voxelmorph,hinkle2018diffeomorphic,chen2022transmorph,kim2022diffusemorph} to project the input images into a latent deformation space, denoted as $\mathcal{Z}$. This space embeds a collection of latent features, $z_{v_0} \in \mathcal{Z}$, associated with the initial velocity field. 

Our GNO is then formulated as an iterative architecture, $z_{v_0} \mapsto z_{v_1} \mapsto \cdots \mapsto z_{v_\tau}$, based on recurrent neural networks. As shown in Fig.~\ref{fig:NetArch}(a), the main architecture of GNO is summarized below:
\begin{itemize}
\item An encoder $P$: lifts the input latent feature, $z_{v_0}$, to a higher dimensional representation with an increased number of channels.
\item A \textit{geodesic evolution layer}: learns a nonlinear mapping function of $z_{v_t} \mapsto z_{v_{t+1}}$ along the geodesic path. 
\item  A decoder $Q$: projects the latent mapping function back to the original input dimension aligned with $z_{v_0}$.
\end{itemize}
Analogous to~\citep{li2020fourier}, the encoder $P$ and decoder $Q$ can be local transformations parameterized by a shallow fully connected neural network. Please note that the functionalities of $P$ and $Q$ slightly differ from a typical image encoder and decoder, as they expand the latent features, $z_0$, into higher-dimensional channels and then project them back to the original channel dimension. This particular design aims to enrich the model capacity and expressivity in the latent space. Next, we will introduce details of the new geodesic evolution layer.  

\paragraph*{\bf Geodesic evolution layer.} We utilize a multilayer neural network with $J$ hidden layers to simulate the geodesic mapping function from $z_{v_{t}}$ to $z_{v_{t+1}}$. To simplify the math notation, we let $u^j$ denote the input of $j$-th hidden layer and $u^{j+1}$ denotes the output, where $j \in \{1, \cdots, J\}$. A composition of a local linear transformation, $W^j$, and a global convolutional kernel, $\mathcal{H}^j$, is used to extract both global and local representations at each hidden layer. Additionally, a nonlinear activation function, $\sigma(\cdot)$, is carefully designed to encourage a smooth evolution of the geodesic path. We update $u^{j} \mapsto u^{j+1}$ by
\begin{align}\label{eq:FreFno}
   u^{j+1} := \sigma(W^j u^{j} + \mathcal{H}^j*u^{j}),
\end{align}
where $*$ represents a convolution operator. 
In order to compute the global convolution in an efficient way, we apply matrix-vector multiplication for two Fourier signals in the frequency domain, and then project it back to the original space.
We employ a Gaussian Error Linear Unit (GELU) \citep{hendrycks2016gaussian} with the smoothing 
operator $K$ to ensure the smoothness of the output signal, we have
$\sigma := K (\text{GELU} (\cdot))$.

Finally, we update $z_{v_{t-1}}$ to $z_{v_t}$ by
\begin{equation*}
z_{v_{t+1}}:= \sigma(W^J, \mathcal{H}^J) \circ \cdots \circ \sigma(W^2, \mathcal{H}^2) \circ \sigma(W^1, \mathcal{H}^1, z_{v_t}),
\label{eq:FneurEPDiff}
\end{equation*}
where $\circ$ denotes a function composition.

\paragraph*{\bf Geodesic loss.} We formulate the geodesic loss function of GNO as an empirical data loss term guided by numerical solutions (e.g., using an Euler integrator) to the geodesic shooting equations. Since our model GNO is now learning a sequence of latent velocity fields $\{z_{v_0}, z_{v_1}, \cdots, z_{v_\tau}\}$, the registration network will decode a time-sequence of $\{v_0, v_1, \cdots, v_\tau\}$ accordingly. 

Given a set of $N$ observations, $\bm{\hat{v}}^n \triangleq \{ \hat{v}_1^n, \cdots, \hat{v}_{\tau}^n\}$, which are numerical solutions to the geodesic shooting equation (Eq.~\eqref{eq:epdiff}) with a given initial condition, we formulate our geodesic loss as the mean squared error (MSE) between the predicted time-dependent velocities and the corresponding velocities computed via the numerical solution to the EPDiff equation, i.e.
\begin{align}
\label{eq:geoloss}
l(\theta_r, \theta_v)  &=  \sum_{n=1}^{N} \| \mathcal{D}_{\theta_v} (\mathcal{G}_{\theta_r}(z_{v_0}))- \bm{\hat{v}}^n \|_2^2 \nonumber \\
&+ \text{Reg}(\theta_r, \theta_v),
\end{align}
where $\mathcal{G}_{\theta_r}(z_{v_0})$ denotes the geodesic mapping function that transforms the latent initial velocity $z_{v_0}$ into a sequence of time-dependent latent velocities ${z_{v_1}, \ldots, z_{v_\tau}}$. The decoder $\mathcal{D}_{\theta_v}(\cdot)$ then maps this latent trajectory back to the image space, reconstructing the deformation process. The term $\text{Reg}(\cdot)$ represents a regularization applied to the network parameters.

\subsection{Joint Learning of GDN}
The loss function of GDN integrates losses from both an unsupervised registration network and the geodesic neural operators informed by real numerical solutions. Given a set of pairwise images $\{S^n, T^n\}_{n=1}^N$, we are now ready to define the joint loss of GDN as
\begin{align}
\label{eq:JointLossFun}
L(\theta_r, \theta_v) &=  \sum_{n=1}^{N} \lambda \, \|(S^n(\phi^{n}_\tau (\theta_v)) - T^n \|^2_2  \nonumber \\
&+ \frac{1}{2} \xi (\mathcal{L} v^n_0 (\theta_v), v^n_0 (\theta_v))  \nonumber \\
&+ \eta \, l(\theta_r, \theta_v) 
\,\,\,\, s.t. \,\, \text{Eq.} ~\eqref{eq:phi_v}, 
\end{align}
where $\lambda$, $\xi$, and $\eta$ are positive weighting parameters to balance the image matching term and the geodesic loss defined in Eq.~\eqref{eq:geoloss}.

\section{Experimental Evaluation}

We demonstrate the effectiveness of our model, GDN, on both 2D synthetic
data and 3D brain MRI scans. Our experimental design focuses on three main perspectives to evaluate the performance of the proposed GDN. We first quantitatively measure the difference between our predicted geodesic path of deformations and the real numerical solutions. We then show the capability of our model in promoting the generalizability and regularizability of the registration network. We finally report the quantitative results of the registration accuracy. Additionally, it is worth noting that the prediction of geodesics designed in GDN bypasses the need for computing numerical solutions of EPDiff in previous learning-based LDDMM registration framework~\citep{yang2017quicksilver,Wang_2020_CVPR, hinkle2018diffeomorphic}, resulting in a reduced testing inference time.

\subsection{Datasets}

\paragraph*{\bf 2D synthetic data.} We demonstrate our model on three synthetic datasets: $2$D synthetic circles with random radius, $2$D hand-written digits MNIST~\citep{lecun1998mnist}, and Google QuickDraw~\citep{jongejan2016quick} - a collection of categorized drawings contributed by online players in a drawing game. For the QuickDraw dataset, we randomly choose $10000$ images that include different classes of envelopes, moons, triangles, shirts, etc. All images were upsampled to the size of $64 \times 64$, and were pre-aligned with affine transformations within each class. \\

\noindent \textbf{3D brain MRI.} We include $828$ T1-weighted 3D brain MRI scans from the Open Access Series of Imaging Studies (OASIS-3) dataset~\citep{Fotenos1032}. This dataset is part of the Learn2Reg challenge~\citep{alsinan2022learn2reg} and has been widely used in the literature as a standard benchmark~\citep{yang2017quicksilver,Wang_2020_CVPR,hinkle2018diffeomorphic,balakrishnan2019voxelmorph,chen2022transmorph}.
The MRI scans were resampled to the dimension of $128 \times 128 \times 128$, with an isotropic resolution of $1.25 \text{mm}^3$. All MRIs have undergone skull-stripping, intensity normalization, bias field correction, and affine alignment. Due to the difficulty of preserving the diffeomorphic property across individual subjects, particularly with large
age variations, we carefully evaluate images from subjects aged $60$ to $90$. We split the dataset into $70\%$ for training, $15\%$ for validation, and $15\%$ for testing. The testing volumes include manually delineated anatomical structures, such as White-Matter (WM), Cerebral-Cortex (CerebralC), Ventricle (Ven), Cerebellum-Cortex (CerebellumC), Thalamus (Tha), Putamen (Puta), Caudate (Caud), Hippocampus (Hipp), and brain stem (Stem).

\subsection{Experimental Design}

\paragraph*{\bf Geodesic learning evaluation.} We evaluate the effectiveness of our proposed GDN in learning geodesic deformations on 2D synthetic data. Specifically, we randomly select $12,000$ image pairs from a combined dataset of MNIST ($2,000$ per digit ranging from $0$ to $4$) and binary circles ($2,000$ pairs). The dataset is partitioned into $70\%$ for training, $15\%$ for validation, and $15\%$ for testing. We compare the predicted velocity fields, $\{v_t\}$, and their associated transformations, $\{\phi_t\}$, with two optimization-based algorithms - (i) the numerical solutions derived from the original geodesic shooting equation (Eq.~\eqref{eq:epdiff}) and (ii) predictions from NeurEPDiff~\citep{wu2023neurepdiff}. We also visualize a sequence of deformed images along the predicted geodesic path. \\

\noindent \textbf{Evaluation on generalizability and regularizability.} To investigate the generalizability and regularizability of GDN in the context of registration, we first train the model on 2D circles and 2D MNIST data (digits 0-4 only) and test it on data from the same dataset but were not seen during training. We then evaluate GDN's performance across different distributions of training and testing data. This includes testing the trained model on the remaining digits (digits 5-9) and a distinct dataset from Google QuickDraw. The final deformed images vs. target images are produced to examine the quality of image alignment. Meanwhile, we visualize the predicted deformations and their determinant of Jacobian maps to examine whether they are well regularized. Additionally, we report the percentage of voxels with a negative Jacobian determinant on the 2D dataset and  the negative values observed in the determinant of Jacobian maps across all methods for real 3D brain MRI datasets.
We compare the performance of GDN with the state-of-the-art deep learning-based diffeomorphic registration methods, notably LagoMorph (LM)~\citep{hinkle2018diffeomorphic}, QuickSilver (QS)~\citep{yang2017quicksilver}, VoxelMorph-diff (VM)~\citep{dalca2019unsupervised}, and TransMorph (TM)~\citep{chen2022transmorph}. Please note that only the LM and QS methods are based on LDDMM. The other two baseline approaches are based on stationary velocity fields and do not necessarily generate geodesic paths unless the space is equipped with a flat affine connections~\citep{lorenzi2013geodesics}. All methods are trained with their best performance reported. \\

\noindent \textbf{Final deformation evaluation.} We evaluate the quality of predicted deformations of GDN by quantitatively measuring its accuracy on real 3D brain MRIs. We perform registration-based segmentation and compare the resulting segmentation accuracy of our GDN with all baseline algorithms using two distinct quantification metrics. One commonly used metric is the dice similarity coefficient (DSC)~\citep{dice1945measures} which evaluates volume overlap between the propagated/deformed segmentation $A$ and the manual segmentation $B$ for each structure. Such a dice score can be computed by DSC$(A, B) = 2(|A| \cap |B|)/(|A| + |B|)$, where $\cap$ denotes an intersection of two regions. \\

\noindent {\bf Statistical evaluation.} We performed paired t-tests~\citep{student1908probable} to compare our proposed model GDN with other benchmarks and reported the corresponding p-values for all models on the DSC scores across different anatomical structures using real 3D brain MRI datasets. This analysis provides insight into whether GDN performs significantly better or differently compared to the baselines in terms of registration accuracy.\\

\noindent \textbf{Parameter setting.} 
The training of all our experiments is implemented on an server with AMD EPYC $7502$ CPU of $126$GB memory and Nvidia GTX $3090$Ti GPUs. 
We train $2$D and $3$D networks using the Adam optimizer~\citep{kingma2014adam} with weight decay as $1e^{-4}$. The $2$D network was trained for $1000$ epochs with a batch size of $100$, while the $3$D network was trained for $2000$ epochs with a batch size of $6$. We used a learning rate of $5e^{-4}$ for our network GDN.
We set the number of integration steps to $\tau = 10$ for the 2D dataset and $\tau = 5$ for the 3D dataset to balance accuracy and efficiency. The smoothing parameters for the operator $K$ in Eq.~\eqref{eq:epdiff} are set as $\alpha = 1.0$, $\beta = 0.5$ for the 2D dataset, and $\alpha = 1.0$, $\beta = 1.0$ for the 3D dataset. The positive weighting parameter for registration is $\lambda = \frac{1}{{0.03}^2}$, with the regularization weight set to $\xi = 2.0$ for 2D and $\xi = 0.15$ for 3D. The number of hidden layers in the GNO is configured as $J = 4$ for 2D and $J = 1$ for 3D. 
The backbone of the registration network is a U-Net architecture with a kernel size of 3. For our GDN, the channel configurations are set to [[8, 16, 16, 8], [8, 16, 16, 16, 8]]. All baseline models, including LM, QS, and VM-diff, use the same U-Net architecture with identical kernel size and channel configurations to ensure a fair comparison. \\

\noindent \textbf{Ablation Study.} We conduct an ablation study on 3D real brain MRIs to evaluate the influence of the number of integration steps ($\tau$) in the geodesic learning module on both registration accuracy and regularity.

Please note that, our model GDN, when excluding the proposed GNO module and geodesic loss, is equivalent to the baseline LagoMorph. This serves as an ablation study to assess the contribution of the GNO and geodesic loss components.

\paragraph*{Numerical scheme in GNO learning.} The numerical solution of the geodesic shooting equation (Eq.~\eqref{eq:epdiff}) that guides the learning of GNO is generated in real-time during the training process. In all experiments, we opt for a commonly employed Euler integrator to numerically solve this shooting equation~\citep{yang2017quicksilver,Wang_2020_CVPR,hinkle2018diffeomorphic,vialard2012diffeomorphic,beg2005computing}. Other advanced integrators, such as Runge-Kutta, can be easily applied.

\subsection{Experimental Results}

\begin{figure*}[!h]
\centering
\includegraphics[width=0.94\textwidth] {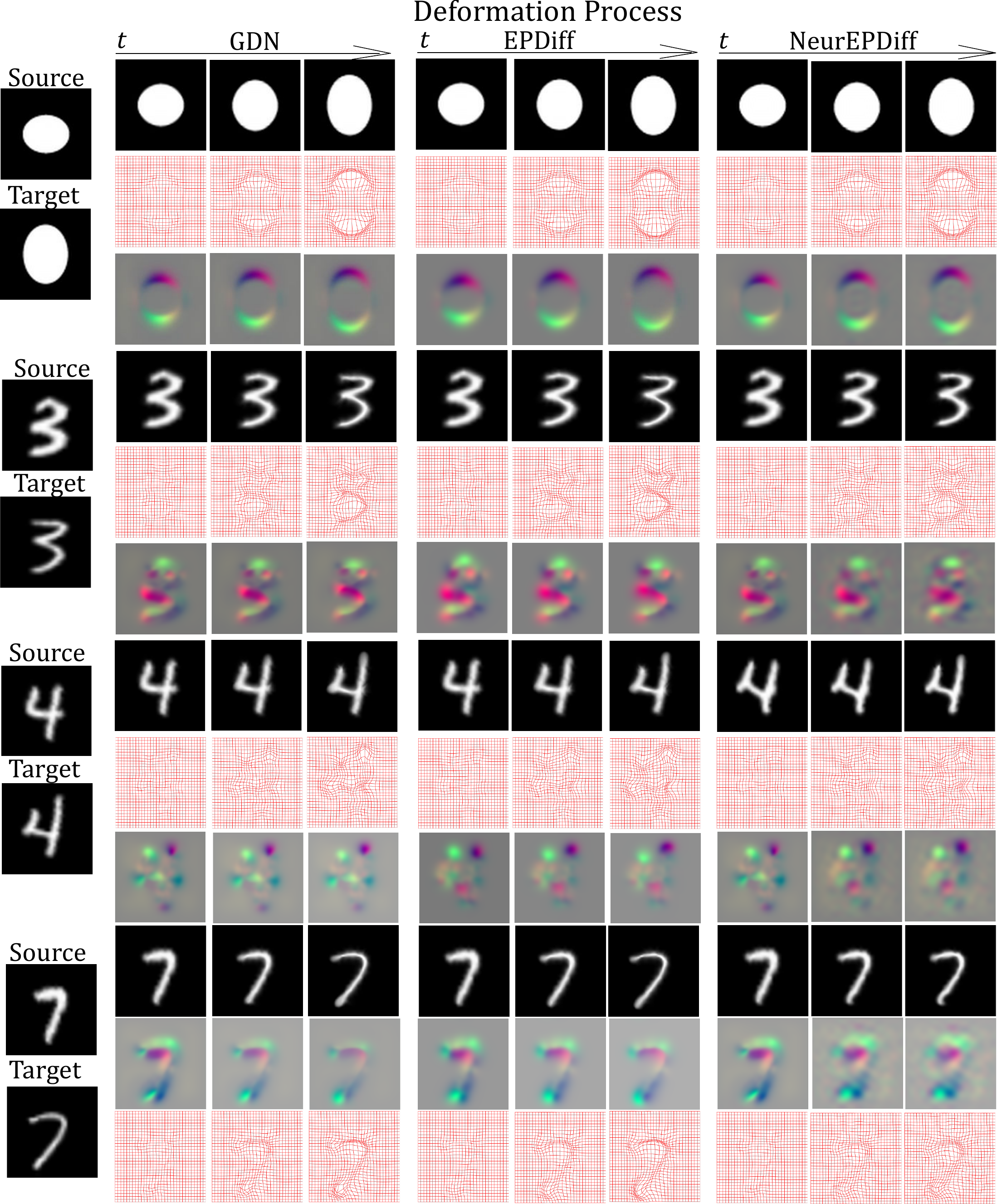}
     \caption{Visulization of predicted geodesics from GDN vs. real numerical solutions~\citep{zhang2015fast} and predictions from NeurEPDiff~\citep{wu2023neurepdiff}. Left to right: source and target images, predicted geodesic deformations along time $t$. Top to bottom: deformed images, transformation fields, and velocity fields.}
\label{fig:conv}
\end{figure*}

In Fig.~\ref{fig:conv}, we present the visualizations of the predicted transformations from GDN alongside numerical solution obtained from two optimization-based algorithms, the original geodesic shooting equation and NeurEPDiff. The deformation processes generated by both GDN and numerical solution show a nice degree of similarity, while GDN slightly outperforms NeurEPDiff.

\begin{figure*}[!t]
\centering
\includegraphics[width=1.0\textwidth] {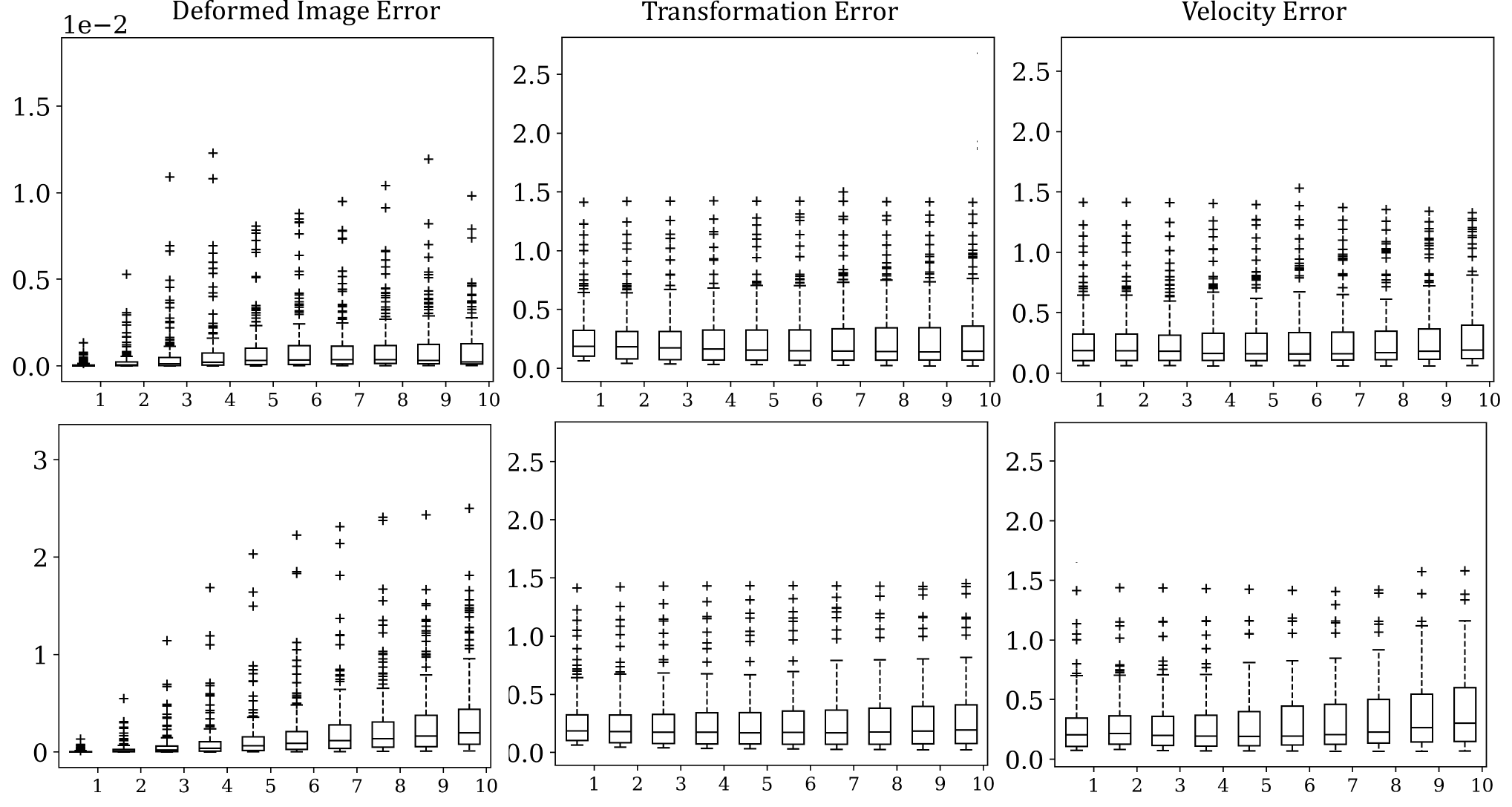}
     \caption{Top to bottom: Comparison of predicted geodesics by GDN/NeurEPDiff vs. numerical integration of EPDiff. Left to right: MSE between deformed images, transformations, and velocities along the geodesic path.}
\label{fig:MSE}
\end{figure*}

Fig.~\ref{fig:MSE} compares the discrepancy between our method and the numerical solution to the EPDiff equation by using NeurEPDiff as a baseline for reference. Specifically, we quantify the mean squared error (MSE) between the predicted time-varying velocity fields and the geodesic velocities obtained through numerical integration of EPDiff. Additionally, we assess the similarity of the resulting transformations and deformed images along the predicted and ground-truth geodesic paths. This result shows that our method, GDN, achieves slightly better performance than NeurEPDiff across these metrics.

\begin{figure*}[!h]
\centering
\includegraphics[width=1.0\textwidth] {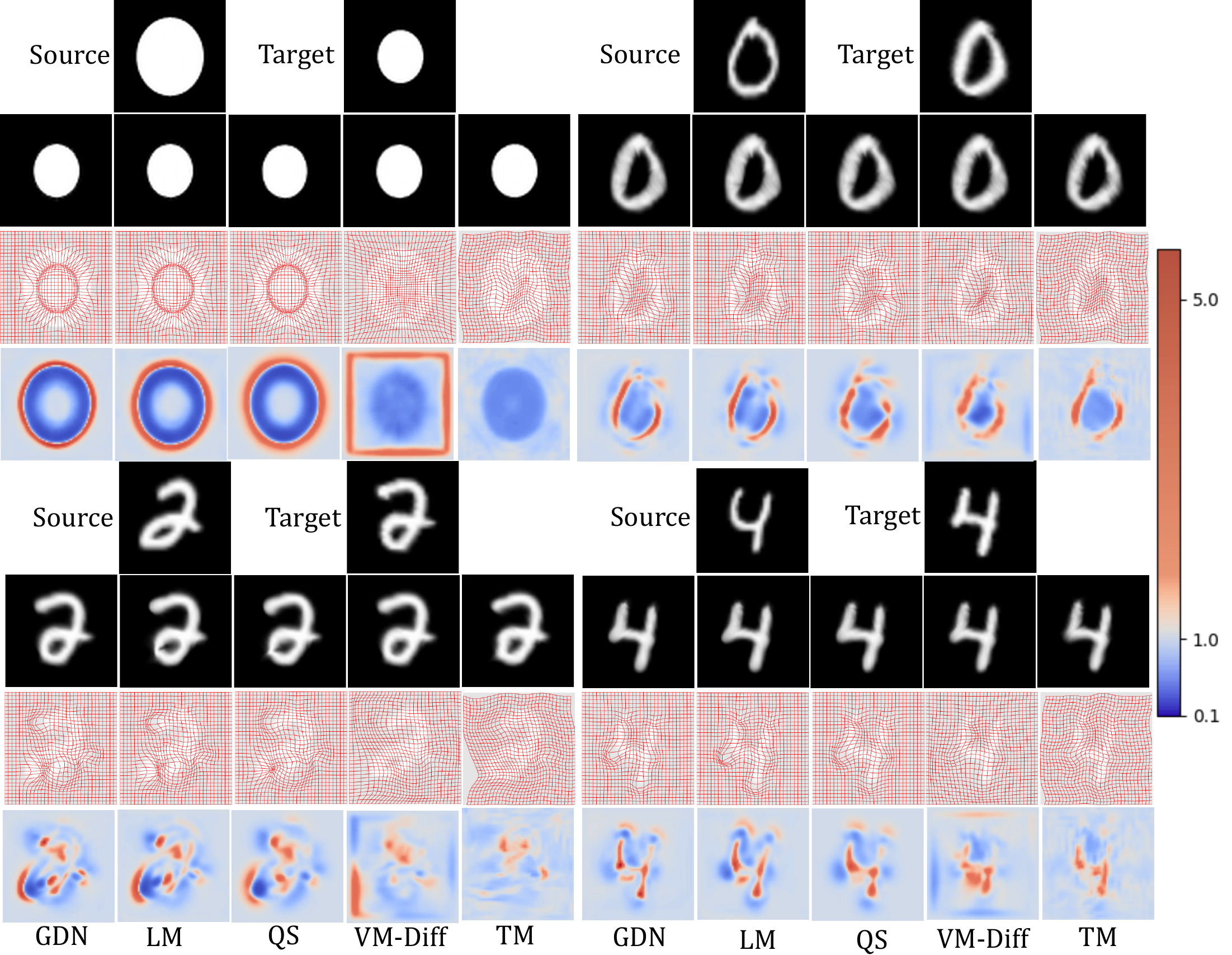}
     \caption{A comparison of predicted transformation grids and their associated determinant of Jacobian maps on in-distribution (ID) testing data from our method GDN vs. baselines. From top to bottom: source and target image pairs, resulting deformed images, predicted transformation grids, and determinant of Jacobian maps. }
\label{fig:withindomain}
\end{figure*}

Fig.~\ref{fig:withindomain} visualizes examples of predicted transformation fields and their associated determinant of Jacobian (DetJac) maps on in-distribution (ID) testing data from all methods. Different values of DetJac indicate different patterns of volume changes. For example, a DetJac value of $1$ indicates no volume change, while DetJac $<1$ reflects volume shrinkage and DetJac $>1$ implies volume expansion. A DetJac value smaller than zero suggests an artifact or singularity in the transformation field, i.e., a failure to regularize the smoothness of deformation fields when the effect of folding and crossing grids occurs. 
While the value of DetJac from all methods do not suggest violations of regularization, our work GDN exhibits smoother transformation grids, indicating more effective regularization. In particular, the estimated transformations from GDN are well regularized to concentrated primarily on the shape of circles or digits. 
Deformation changes on background areas (particularly with no intensity changes between images) have little-to-zero values. 

\begin{figure*}[!ht]
\centering
\includegraphics[width=0.95\textwidth] {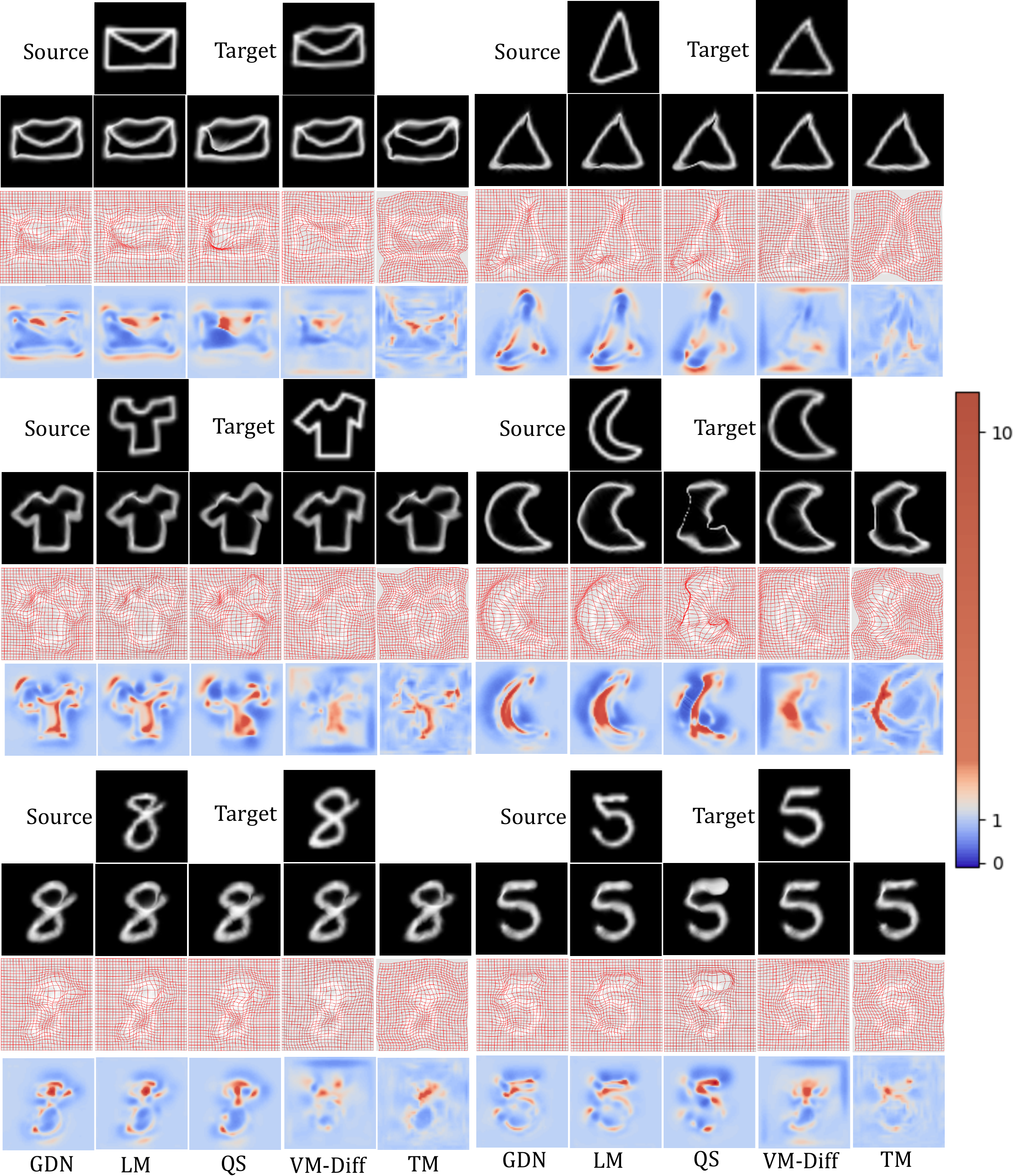}
     \caption{A comparison of predicted transformation grids and their associated determinant of Jacobian maps on out-of-distribution (OOD) testing data from our method GDN vs. baselines. Top to bottom: source and target images, deformed source images, predicted deformation fields, and determinant of Jacobian maps.}
\label{fig:ood}
\end{figure*}

Fig.~\ref{fig:ood} displays a comparison of predicted transformation fields generated from out-of-distribution testing (OOD) data. The results indicate that GDN consistently produces smoother transformation grids, ensuring well-aligned source-to-target images. These consistent findings suggest that our introduced geodesic learning approach in image registration contributes to an enhanced network generalizability and regularizability compared to the other baseline methods. 

\begin{figure*}[!h]
\centering
\includegraphics[width=0.95\textwidth] {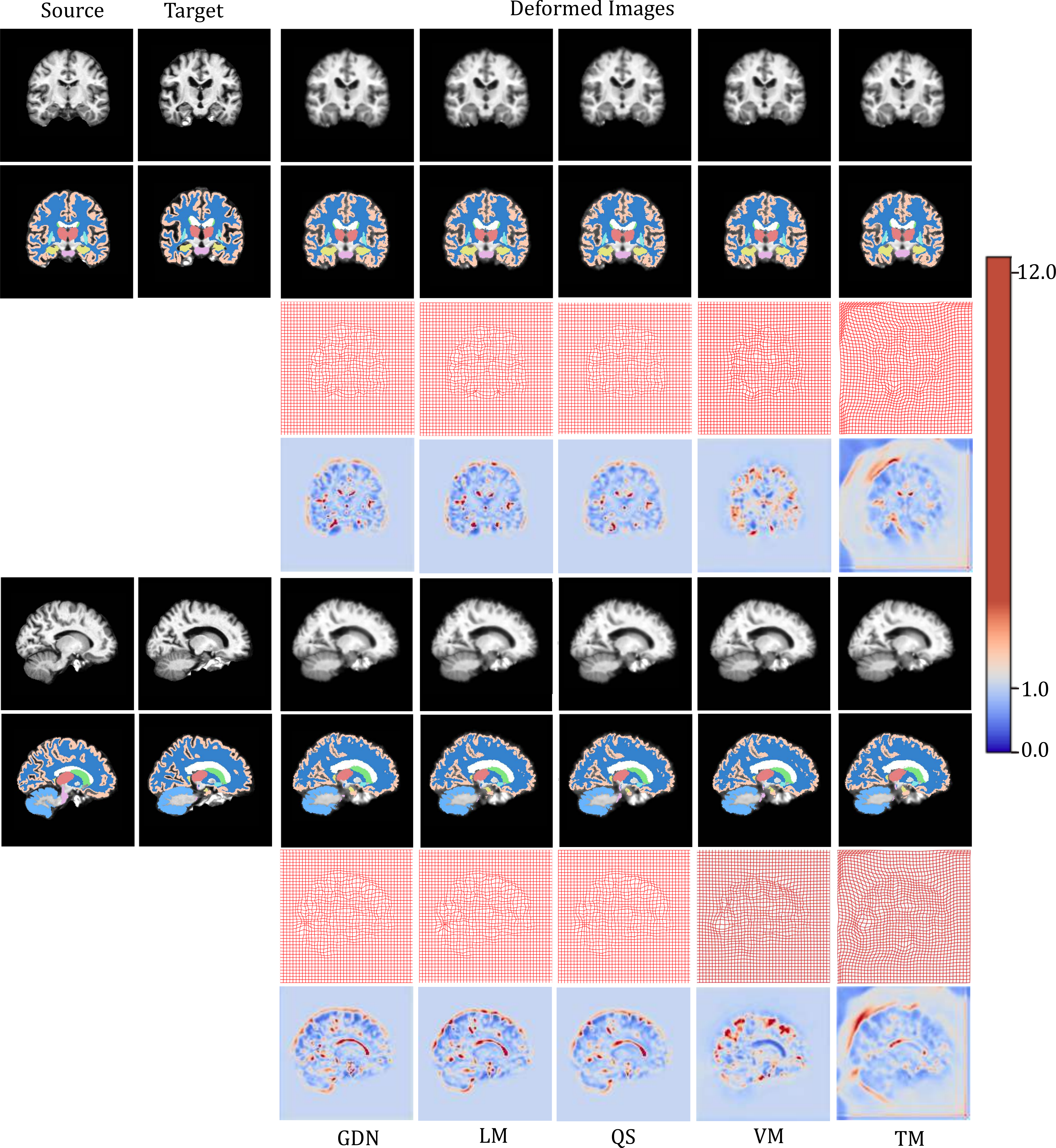}
     \caption{Top to bottom: examples of 3D brain MRIs (demonstrated in coronal and sagittal views), registration-based segmentation maps overlaid on brain MRIs, predicted deformation fields, and determinants of Jacobian maps. Left to right: source and target images; deformed source images (segmentation
     labels) across all methods.}
\label{fig:seg}
\end{figure*}

Fig.~\ref{fig:seg} presents visual comparisons of deformed images, registration-based segmentation maps overlaid on 3D brain MRIs, predicted deformation fields, and determinants of Jacobian maps for all methods, shown in both sagittal and coronal views. Our method demonstrates a similar smoothing effect on transformation grids with LM and QS, while more clearly highlighting biologically plausible growth patterns, such as ventricular enlargement (highlighted in red in the determinants of Jacobian maps). These patterns are more interpretable than those produced by the baseline methods, VM and TM.

\begin{figure*}[!t]
\centering
\includegraphics[width=0.95\textwidth] {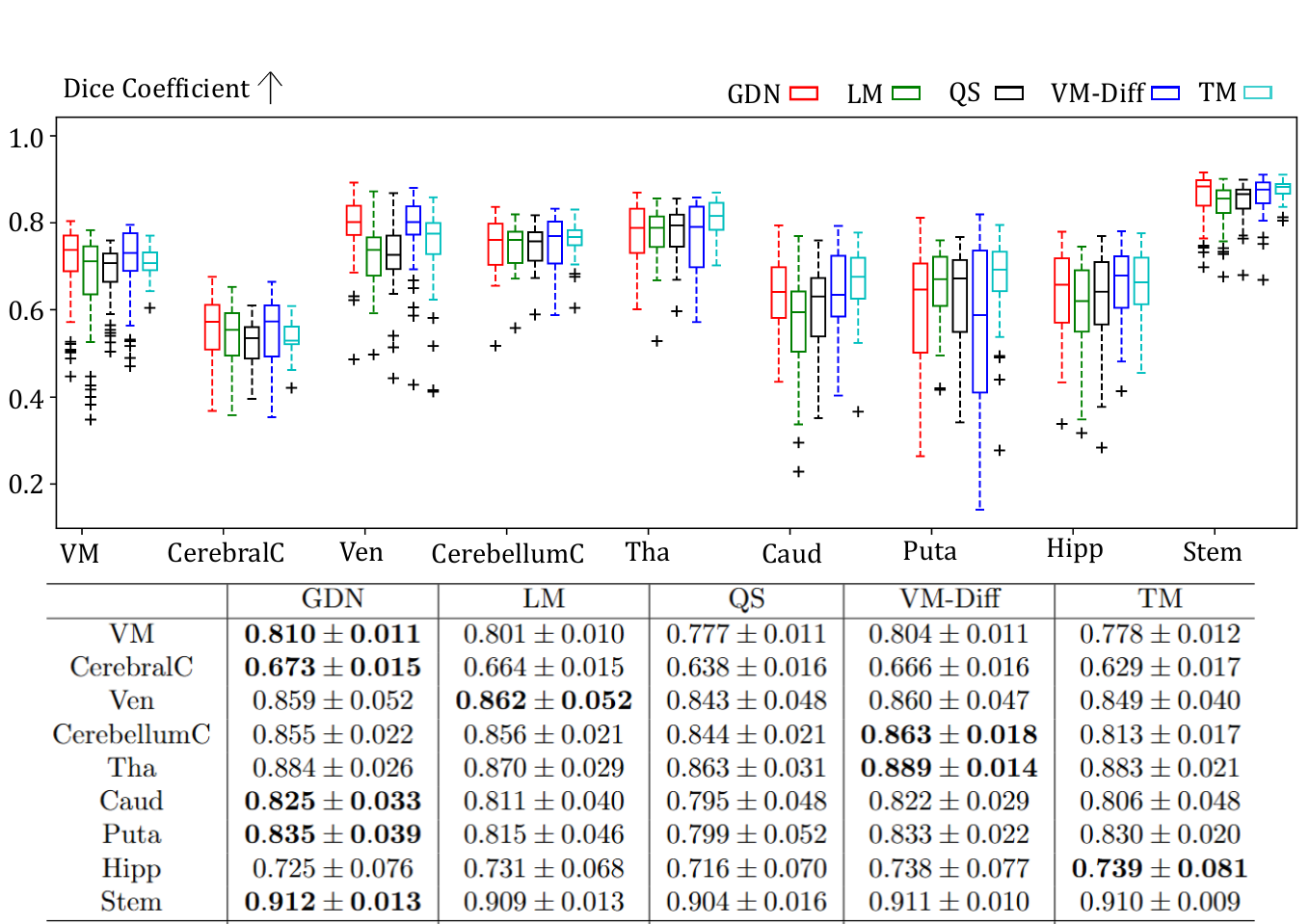}
     \caption{Top: A comparison of dice score by propagating the deformation field to the segmentation labels on nine brain structures White-Matter (WM), Cerebral-Cortex (CerebralC), Ventricle (Ven), Cerebellum-Cortex (CerebellumC), Thalamus (Tha), Putamen (Puta), Caudate (Caud), Hippocampus (Hipp), and brain stem (Stem).}
\label{fig:braindice}
\end{figure*}

Fig.~\ref{fig:braindice} displays a quantitative comparison between the manually labeled vs. propagated segmentation labels deformed by deformations fields predicted from all methods. The top panel reports the statistics of dice scores (the higher the better) over eight brain structures from hundreds of testing registration pairs. The bottom panel summarizes the corresponding means and standard deviations. It shows that our method GDN is able to produce superior-quality final deformation fields in comparison with other baselines. Overall, our GDN achieves notably higher Dice on average. Meanwhile, an additional benefit of GDN is to forgo the computation of the geodesic shooting equation inherent in conventional LDDMM-based registration networks. This step can substantially increase computational cost in the testing phase, particularly when high-dimensional images. In this experiment, GDN demonstrates its capability to predict geodesics in {\bf $0.026s$}, outperforming the other two baselines in the context of LDDMM - LM and QS ($0.036$ seconds). It's important to note that the other two baselines - VM-diff~\citep{balakrishnan2019voxelmorph} and TM~\citep{chen2022transmorph} models employ stationary velocity fields to parameterize the deformation fields, treating the velocities as constant over time. 

We also report the corresponding p-values from pairwise statistical tests comparing our proposed model, GDN, with four baseline methods across the anatomical brain structures based on Dice scores. The p-values for GDN vs. LM ($<0.001$), QS ($<0.001$), VM ($<0.001$), and TM ($<0.001$) are all well below the significance threshold of $0.05$, indicating that GDN performs significantly differently compared with baselines.

\begin{figure*}[!h]
\centering
\includegraphics[width=0.6\textwidth] {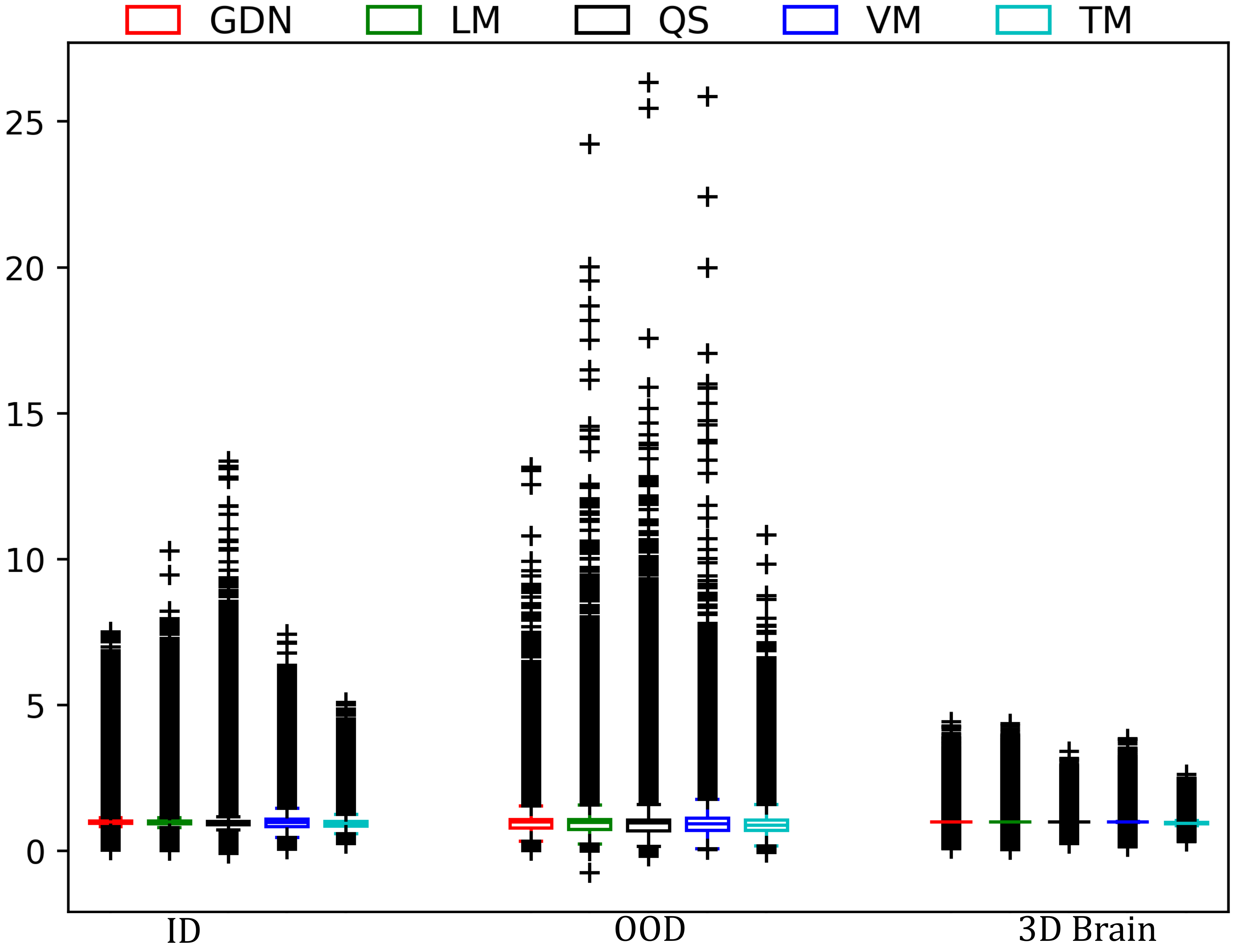}
     \caption{A report on the negative values in determinant of Jacobian of the transformations on three different datasets: in-distribution (ID) testing data, out-of-distribution (OOD) testing data, and real 3D brain MRIs.}
\label{fig:negJac}
\end{figure*}

Fig.~\ref{fig:negJac} presents a quantitative comparison of negative values in the determinant of Jacobian(DetJac) of the transformations across three different datasets --- in-distribution (ID) testing data, out-of-distribution (OOD) testing data, and real 3D brain MRIs. 
The results indicate that for both the ID testing data and the 3D brain MRIs, there are barely violations of regularization for all methods. However, it is worth noting that our method, GDN, demonstrates advantages on the OOD testing data, with zero negative values reported. However, when tested on data outside the training distribution, the percentage of negative Jacobian determinants begins to diverge across models: GDN (0.005\%), LM (0.132\%), QS (0.255\%), VM-Diff (0.038\%), and TM (0.011\%). These findings indicate the potential of GDN to achieve improved generalizability and regularization compared to other baselines.






 


\begin{table}[ht]
\centering
\small
\setlength{\tabcolsep}{3pt}
\renewcommand{\arraystretch}{1.5}   

\begin{tabular}{>{\centering\arraybackslash}m{2.1cm}|ccc|cc}
\hline
  & \multicolumn{3}{c|}{LDDMM-Based} & \multicolumn{2}{c}{SVF-Based} \\ \cline{2-6}
  & GDN & LM & QS & VM\text{-}Diff & TM \\ \hline

\tallerow\shortstack{Number of\\Parameters}
  & \tallerow 79.3K & \tallerow 56.5K & \tallerow 56.5K
  & \tallerow 56.5K & \tallerow 46.5M \\ \hline

\tallerow\shortstack{Training\\(per epoch)}
  & \tallerow 61.5s & \tallerow 14.61s & \tallerow 20.5s
  & \tallerow 20.78s & \tallerow 33.39s \\ \hline

\tallerow\shortstack{Testing}
  & \tallerow 0.030s & \tallerow 0.043s & \tallerow 0.043s
  & \tallerow 0.017s & \tallerow 0.020s \\ \hline
\end{tabular}

\caption{A summary of model parameters, training time per epoch,
and testing time on 3D brain MRI data.}
\label{tab:ModelSize}
\end{table}

Tab.~\ref{tab:ModelSize} summarizes the model parameters, training time per epoch, and inference time for GDN versus the baselines. Our model, GDN, has a larger number of parameters than LM and QS (under the same LDDMM setting) due to the added geodesic learning module. In addition, training GDN requires solving the EPDiff equation numerically to supervise the geodesic loss. These two factors contribute to GDN's longer training time per epoch compared to LM and QS. However, a key advantage of GDN lies in inference: once trained, GDN does not require numerical integration during testing. As a result, GDN achieves faster inference time (0.030s) than LM and QS (0.043s), and we expect this advantage to become more pronounced with higher-dimensional image data. When compared to VM-diff and TM under the stationary velocity field setting, GDN does not show a significant advantage in training or inference time, primarily due to its larger model size relative to VM-diff. The model TM, while larger in size, benefits from parallel computation and hence is more time-efficient.

\begin{figure*}[!h]
\centering
\includegraphics[width=0.5\textwidth] {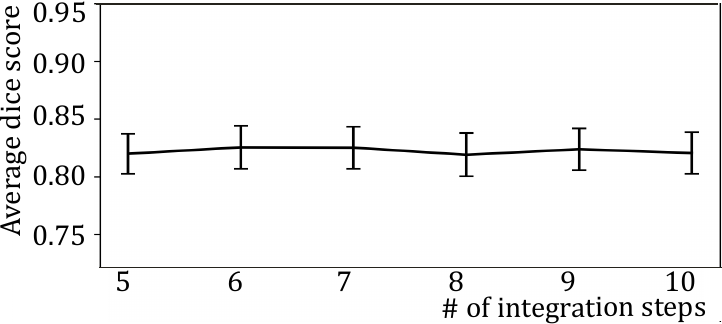}
      \caption{A comparison of average dice score on 3D brain MRIs with a varying number of integration steps.}
\label{fig:ablation}
\end{figure*}

Fig.~\ref{fig:ablation} presents a quantitative comparison of the average dice scores on 3D brain MRIs with different integration steps of our proposed module GNO. The results show that our method achieves stable dice accuracy based on the deformed template segmentation labels as the number of integration steps increases from five to ten.

\section{Conclusions}
This paper presents a novel deep network, named as GDN, that {\em for the first time learns geodesics of deformation spaces in learning-based image registration methods}. In contrast to current approaches that solely fit the registration loss, we develop a geodesic deformation subnetwork guided by the fundamental physical processes of deforming images from one to another. To achieve this, we treat the geodesics as unknown mapping functions directly learned from the data via a carefully developed neural operator. A composition of integral operators and smooth activation functions is then formulated to effectively approximate such mappings. Our proposed GDN jointly optimizes a newly defined geodesic loss function with an alternating optimization scheme. Experimental results on both 2D synthetic data and 3D brain MRI scans show that our model gains an improved generalizability with more effective regularizations on the final deformation fields. 

Our work GDN is an initial step towards quantifying and analyzing geometric deformations from images within deep neural networks. The proposed geodesic learning lays a foundation for performing further quantitative and interpretable analysis in the latent spaces of deformation fields. Our potential future works may focus on learning geodesic regression and population-based variability in deep nets. 
\acks{This work was supported by NSF CAREER Grant 2239977.}

\ethics{The work follows appropriate ethical standards in conducting research and writing the manuscript, following all applicable laws and regulations regarding treatment of animals or human subjects.}

\coi{We declare we do not have conflicts of interest.}

\data{The Brain MRIs used in training and testing in this paper are from the Open Access Series of Imaging Studies (OASIS-3) dataset~\citep{lamontagne2019oasis}, which are readily accessible and user-friendly. 
Readers interested in evaluating the accuracy of the method can access and utilize the publicly available datasets. The code for this study is also publicly available at \url{https://github.com/nellie689/GDN}, with an archived version on Zenodo at \url{https://zenodo.org/records/17733235}.}

\bibliography{main}





\end{document}